\documentclass[10pt,conference]{IEEEtran}
\IEEEoverridecommandlockouts

\usepackage{cite}
\usepackage{amsmath,amssymb,amsfonts}
\usepackage{algorithmic}
\usepackage{array}
\usepackage{graphicx}
\usepackage{subfigure}
\usepackage{textcomp}
\usepackage{xcolor}
\usepackage{gensymb}
\usepackage{hyperref}
\usepackage{csquotes}
\usepackage{eso-pic}

 \AddToShipoutPicture*{\small \sffamily\raisebox{1.2cm}{\hspace{1.8cm}978-1-7281-2009-6/\$31.00@2019 IEEE}}

\makeatletter
\newcommand{\printfnsymbol}[1]{%
  \textsuperscript{\@fnsymbol{#1}}%
}
\makeatother
\graphicspath{ {images/} }

\def\BibTeX{{\rm B\kern-.05em{\sc i\kern-.025em b}\kern-.08em
    T\kern-.1667em\lower.7ex\hbox{E}\kern-.125emX}}
\begin{document}

\title{Transfer Learning Using Ensemble Neural Networks for Organic Solar Cell Screening \\
}

\author{\IEEEauthorblockN{
Arindam Paul\thanks{\printfnsymbol{1}equal contribution}\IEEEauthorrefmark{1},
Dipendra Jha\printfnsymbol{1},
Reda Al-Bahrani,
Wei-keng Liao,
Alok Choudhary and
Ankit Agrawal
}

%\author{\IEEEauthorblockN{
%Arindam Paul,
%Dipendra Jha,
%Reda Al-Bahrani,
%Wei-keng Liao,
%Alok Choudhary and
%Ankit Agrawal
%}

\IEEEauthorblockA{Electrical Engineering and Computer Science, Northwestern University}

{Email: { arindam.paul, dkj755, rav650, wkliao, choudhar, ankitag}@eecs.northwestern.edu}\\}

\maketitle

\begin{abstract}
Organic Solar Cells are a promising technology for solving the clean energy crisis in the world. However, generating candidate chemical compounds for solar cells is a time-consuming process requiring thousands of hours of laboratory analysis. For a solar cell, the most important property is the power conversion efficiency which is dependent on the highest occupied molecular orbitals (HOMO) values of the donor molecules.
Recently, machine learning techniques have proved to be very useful in building predictive models for HOMO values of donor structures of Organic Photovoltaic Cells (OPVs). Since experimental datasets are limited in size, current machine learning models are trained on data derived from calculations based on density functional theory (DFT). Molecular line notations such as SMILES or InChI are popular input representations for describing the molecular structure of donor molecules. The two types of line representations encode different information, such as SMILES defines the bond types while InChi defines protonation. In this work, we present an ensemble deep neural network architecture, called SINet, which harnesses both the SMILES and InChI molecular representations to predict HOMO values and leverage the potential of transfer learning from a sizeable DFT-computed dataset- Harvard CEP to build more robust predictive models for relatively smaller HOPV datasets. Harvard CEP dataset contains molecular structures and properties for 2.3 million candidate donor structures for OPV while HOPV contains DFT-computed and experimental values of 350 and 243 molecules respectively. Our results demonstrate significant performance improvement from the use of transfer learning and leveraging both molecular representations.

\end{abstract}

\section{Introduction}
Based on current statistics, energy consumption had increased from 238 Exajoules (EJ) in 1972 to 464 EJ in 2004. A further 65\% increase is projected by the year 2030~\cite{stocker2014climate}. Sustained usage of fossil fuels leads to irreversible changes to the planet with sea level rising from 1.7 to 3.2 mm per year and ocean temperatures increasing~\cite{hoffman2015computational,stocker2014climate}. It is imperative to search for versatile and cost-efficient clean energy solutions to prevent further irreversible damage. One limitation with renewable energy is that it is difficult to generate the quantities of electricity that are as large as those produced by traditional fossil-fuel generators~\cite{boyle1997renewable}. Wind and hydro-energy solutions require expensive installations~\cite{turner1999realizable} and maintenance, which requires large government grants, and are dependent on weather and climate conditions~\cite{baxter2005biomass}.
Moreover, most renewable energy technology is new and has enormous capital costs compared to traditional fossil fuels. Solar energy provides a more cost-effective solution with faster installation, and more predictive energy outputs based on the Bureau of Meteorology and National Aeronautics and Space (NASA) reports~\cite{NASAClean}. Although inorganic silicon-based solar energy systems are currently more conventional, organic or plastic photovoltaic (OPV)~\cite{yu2014towards} technology has become very popular because of its flexibility. Organic or Plastic Technology is very versatile as demonstrated by how plastics in consumer goods can be made very hard and durable, or very light or transparent as dictated by needs~\cite{brabec2011organic}. Further,  manufacturing costs are lower for organic solar cells compared to silicon-based materials due to the ease of device manufacturing, and lower cost of organic components compared to silicon~\cite{abdulrazzaq2013organic}.

However, the main bottleneck in the deployment of organic solar cells is that the search for candidate chemical compounds for creating organic solar cells is very time consuming~\cite{forrest2005limits}; it can take up to thousands of hours of laboratory analysis. For a solar cell, the most important property is power conversion efficiency (PCE) or the percentage of electricity which can be generated due to the interaction of electron donors and acceptors after absorption of energy from the sun. The PCE is dependent on the highest occupied molecular orbital (HOMO) energy of the donor and the lowest unoccupied molecular orbital (LUMO) energy of the acceptor molecule~\cite{scharber2006design}. However, as the LUMO values across known acceptors do not vary much, and only a few acceptor molecules exist, predicting HOMO values of donor molecules can give us estimates of PCE when those donors are used in solar cells. 
% The Scharber model~\cite{scharber2006design} provides a relation between the voltage $V_{oc}$ and the energies of the HOMO and the LUMO level of the donor and acceptor molecules, which in turn can be related to the power conversion efficiency (PCE), the maximum efficiency of solar cells. 
% In the following equation, $J_{sc}$ is the short-circuit current density, FF is electrical fill factor and $P_{in}$ is incident-light intensity. $E^{Donor}HOMO$ and $E^{Acceptor}LUMO$ indicate the HOMO and LUMO energy levels of the donor and acceptor molecules respectively. 
%  \[ V_{oc} = (1/e)(E^{Donor}HOMO-E^{Acceptor}LUMO)-0.3V \]
% \[ PCE = 100*(V_{oc}*FF*J_{sc})/P_{in} \] 

There are two main issues with the current practice of building predictive models using machine learning (ML) techniques. First, these predictive models are built using a single representation of the molecular structure - line notations~\cite{warr2011representation} such as SMILES or InChI, molecular fingerprints~\cite{pyzer2015learning} or molecular graphs~\cite{duvenaud2015convolutional}. Line notations are increasingly becoming popular for use in ML models as molecular fingerprints are difficult to interpret and models trained on molecular graphs usually perform worse~\cite{goh2017smiles2vec}. However, these approaches have restricted themselves to only one type of line notation -- either SMILES or InChI as input representations for predictive models. This limits the information that can be harnessed from these representations as SMILES and InChI express the molecular structure in very different ways. SMILES defines the chemical bond types present in the molecular structure from which one can infer the protonation while InChI serves the opposite purpose- it defines the protonation from which one can infer the chemical bond types present in the molecular structure.
Also, SMILES was designed to be read and written by humans whereas InChi was intended to ignore tautomeric form and be more consistent. Since the two line representations are distinct in their properties,
a predictive model can benefit from the use of both of them.
Second, most of the datasets, especially, experimental datasets are limited in size.
Hence, current ML models are either built using publicly available large DFT-computed datasets such as the Harvard CEP dataset~\cite{pyzer2015learning, CEPData} or other limited experimental or DFT-computed datasets which are relatively smaller in size and hence, the model cannot learn the required data representation for making robust predictions. 
In this work, our goal is to leverage together larger DFT-computed datasets with relatively smaller datasets such as experimental observations and combine both types of line representations- SMILES and InChI -- to build more robust predictive models for predicting HOMO values for donor candidates for OPV.

We present an ensemble deep neural network architecture, called SINet, which leverages both the SMILES and InChI molecular representations to learn to predict the HOMO values, and leverage transfer learning from large datasets to build more robust predictive models for a relatively smaller dataset.
SINet is composed of two identical branches for both types of inputs; each branch consists of 1-D CNN layers followed by LSTM layers.
The features learned by the two branches from the two input representations are combined and fed into a fully connected network for predicting the regression output of HOMO value.
The deep neural network architecture of SINet enables us to perform transfer learning from a large dataset to relatively smaller dataset in a similar domain. Transfer learning has already been adopted in the fields of computer vision, natural language processing and other application domains~\cite{pan2010survey, hoo2016deep}.

Our source dataset for transfer learning is the Harvard CEP dataset~\cite{pyzer2015learning, CEPData} which contains molecular structures and properties for 2.3 million candidate donor structures for OPV.
For the target dataset, we leverage DFT-computed and experimental values of 350 and 243 molecules respectively, from the HOPV~\cite{TheHarva43:online}.
Our results demonstrate significant performance improvement from the use of both types of inputs- the MAPE drops from 0.972\% and 0.457\% using SMILES and InChI respectively, to 0.213\% when using them together using the SINet architecture.
We also find significant benefit from using transfer learning from Harvard CEP to the HOPV datasets.
The MAPE for experimental and DFT-computed datasets from HOPV drops from 2.782\% to 1.513\% and 2.118\% to 1.478\%, respectively.
Since the model is first trained on a large dataset, it learns the required set of features from the input data representation, and this helps in learning the similar features present in the smaller target dataset, on which the model is fine-tuned. Our results demonstrate significant benefit from the use of both types of input representations as well as from transfer learning from a larger dataset. It showcases that leveraging machine learning with computational and experimental chemistry can play an essential role in the expedition of a systematic design of high-efficiency OPV materials, and holds significant promise as a potential solution to future energy needs.
\section{Background and Related Works}
In this section, we present a description of the OPV technology and Scharber model, and the two molecular representations we use in this work -- SMILES and InChI, and discuss existing machine learning systems for predictive modeling for materials. 

\subsection{Organic Photovoltaic Cells}

\begin{figure}[h]
\centering
\includegraphics[width = 0.5\textwidth, keepaspectratio=true]{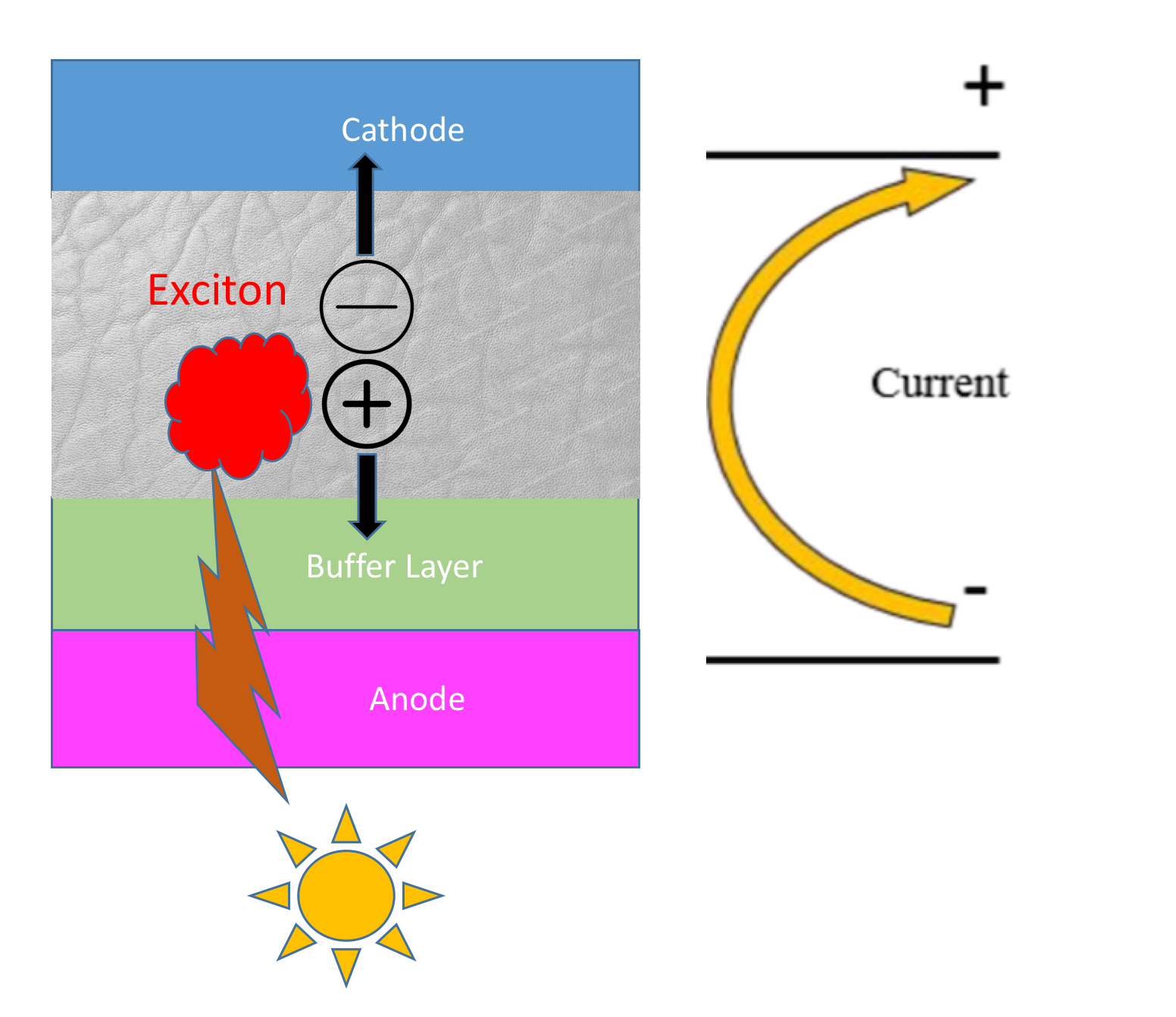}
\caption{Photo-electricity generation in a bulk heterojunction Organic photovoltaic cell~\cite{kanal2013efficient}. When the photons from the sun hit the surface of the OPV device, an electron from the donor is excited and combines with the corresponding hole at the acceptor layer to form an exciton. Electricity is generated when the exciton splits at the interface, and electrons move from the cathode to the anode. }
\label{fig:OPV}
\end{figure}

Among current solar cell design paradigms, organic photovoltaic cell technology is a promising technology for the inexpensive and versatile utilization of solar energy. The traditional development of new OPV materials is predominantly based on empirical intuition or experience of materials scientists. A new design idea is followed by a labor-intensive synthesis, characterization, and prototype device optimization.  Hence, the problem space of OPV renewable energy research is notably complicated as the design of successful OPV materials is a multifaceted problem. The conversion of sunlight into electricity can be achieved using a solar cell and is one of the most attractive future sources of energy. Ever since the development of the first solar cells, there has been an accelerated and comprehensive exploration for cost-effective photovoltaics. OPVs are potential cost-effective and lightweight alternatives to silicon-based solar cells and could lead to the most substantial reduction of production cost. After being excited with light, firmly bound electron-hole pairs (excitons) are generated. As illustrated in Figure~\ref{fig:OPV}, an OPV works by absorbing a photon emitted by the sun. The photon carries energy that is used to excite an electron off a donor layer, often comprised of a semiconducting polymer. 

However, for the solar cell to generate electricity, the electron and hole must be separated and subsequently collected at electrodes of opposite polarity. In order to accomplish this, the exciton bond must be broken.  This happens at the donor-acceptor interface, where the exciton splits into separate free electron and hole. As the charges separate further, they can reach electrodes which upon becoming charged, generate electricity as the electrons move from the cathode to the anode.  

OPVs have the advantage of combining the versatility and flexibility of plastics with photo-electronics. They can be made semi-transparent, and moldable into different forms and shapes. Researchers have even tried spray-coating OPVs on various surfaces~\cite{hoth2009topographical}. Nonetheless, production scale small-area devices yield efficiency of only about 5\% ~\cite{xue2004asymmetric}, with laboratory experiments yielding the highest efficiency of around 10\%, and hence there is much room for improvement. Several models predict the efficiencies to reach 15\%  assuming the usage of state of the art materials and device architectures~\cite{scharber2013efficiency}. 
The conventional process for the generation of such devices is iterative and time-consuming. However, due to the labor-intensive process of generating candidates for OPVs, producing virtual screening techniques as elucidated by Pyzer-Knapp et al.~\cite{pyzer2015learning} and our current study can potentially fasten the process considerably. 

\subsection{Scharber Model}

\label{subsec:scharber}

 %%paraphrase 
 For a solar cell, the most important property is power conversion efficiency or the amount of electricity which can be generated due to the interaction of electron donors and acceptors. The Scharber model~\cite{scharber2006design} provides a relation between the voltage $V_{oc}$ and the energies of the HOMO and the LUMO level of the donor and acceptor molecules, which in turn can be related to the power conversion efficiency (PCE), the maximum efficiency of solar cells. In the following equation, $J_{sc}$ is the short-circuit current density, FF is electrical fill factor and $P_{in}$ is incident-light intensity. $E^{Donor}HOMO$ and $E^{Acceptor}LUMO$ indicate the HOMO and LUMO energy levels of the donor and acceptor molecules respectively. 
 \[ V_{oc} = (1/e)(E^{Donor}HOMO-E^{Acceptor}LUMO)-0.3V \]
\vspace{-0.5cm}
\[ PCE = 100*(V_{oc}*FF*J_{sc})/P_{in} \] 

However, it is necessary to assert that the predictions from the Scharber analysis are restrained by various assumptions and the quality of the input data. The assumptions include requirements related to complicated bulk and interface behavior, and exciton and charge carrier dynamics. Hence, the resultant values should be understood as the potential performance that can be achieved if the assumptions are met. 

\subsection{SMILES}
Line notations are linear representations of chemical structures which encode the connection table and the stereochemistry of a molecule as a line of text ~\cite{warr2011representation}. Simplified Molecular Input Line Entry System (SMILES)~\cite{weininger1988smiles}, ~\cite{toolkit1997daylight} is the most popular specification in the form of a line notation for describing the structure of chemical species using short ASCII strings encoding molecular structures and specific instances. One or more organic molecules attach to form long continuous chains known as branches. In the SMILES notation, branches are described by parentheses.

Molecules and reactions can be specified using ASCII characters representing atom and bond symbols. Most molecular editors and libraries can import SMILES strings. The other advantages of the SMILES format are that it is both human readable and writable. Further, it encodes the stereochemistry of the molecule intuitively. 
In this work, we limit ourselves to character level representation and do not explicitly encode the grammar. 
Based on the canonical ordering of atoms,  unique SMILES is generated by depth-first search.

% %In Figure~\ref{fig:SMILES}, we observe that for each branch, there is a parentheses sign. 
% The presence of aromaticity (a property which makes an organic molecule more stable provided it is planar or all atoms lie on the same plane) is depicted in lower case. In Figure~\ref{fig:SMILES}, regular carbon and sulfur are represented as c and s respectively and non-aromatic bonds are represented in upper case (C, S and O representing carbon, sulfur and oxygen).

\subsection{InChI}
The IUPAC International Chemical Identifier (InChI)~\cite{heller2015inchi} was developed by IUPAC and  NIST (National Institute of Standards and Technology). 
It is a textual identifier for chemical substances which provides a standard way to encode molecular information. Every InChI string starts with the string \enquote{InChI=} followed by the version number and letter S in the case of standardized InChIs. Rest of the InChI string is structured as a sequence of layers and sub-layers. Each layer provides a specific type of information, and are separated by \enquote{/}. The InChI algorithm transforms the structural information of the molecule into a unique InChI identifier in a three-step process. The first step is normalization which removes redundant information. This is followed by canonicalization that generates a unique number label for each atom. The last step is serialization that produces a string of characters. 

\begin{table*}[h!]
  \centering
   \caption{Examples of set of similar chemical compounds with their corresponding SMILES and InChI notations with explanation}
   \label{tab:example_compounds}
  \begin{tabular}{ | c | c | m{10cm} | }
    \hline
    Compound 1 &  Compound 2 & Line Notations (with explanation)\\ \hline
    %Start of first eg  
    \begin{minipage}{.2\textwidth}  
      \includegraphics[width=\linewidth, height=20mm]{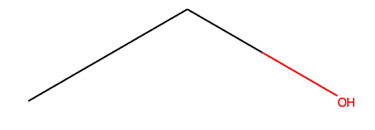}
      \center Ethanol
    \end{minipage}
    &
    \begin{minipage}{.2\textwidth}
      \includegraphics[width=\linewidth, height=20mm]{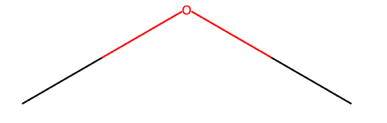}
      \center Dimethyl Ether
    \end{minipage}
    & 
      \begin{itemize}
        \item SMILES: CCO and COC
        \item InChI: InChI=1S/C2H6O/c1-2-3/h3H,2H2,1H3 and  InChI=1S/C2H6O/c1-3-2/h1-2H3
        \item \enquote{C2H6O} means that the first and second atoms (1 and 2) are C atoms and the third (3) is an O atom.  The connectivity is 1-2-3 for ethanol and 1-3-2 for dimethyl ether.  For ethanol atom 3 has 1 H atom, atom 2 has 2 H atoms, and atom 1 has 3 H atoms.  For dimethyl ether atom 1-2 have 3 H atoms, while atom 3 has none. 
      \end{itemize}
    \\ \hline
    %Start of second eg
      \begin{minipage}{.2\textwidth}
      \includegraphics[width=\linewidth, height=20mm]{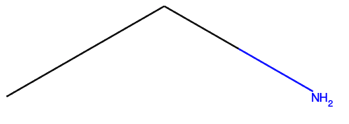}
       \center Ethylamine
    \end{minipage}
    &
    \begin{minipage}{.2\textwidth}
      \includegraphics[width=\linewidth, height=20mm]{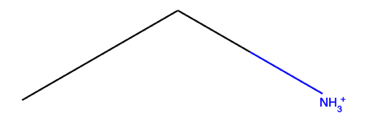}
      \center Ethylammonium
    \end{minipage}
    & 
  \begin{itemize}
        \item SMILES: CCN and CC[NH3+]
        \item InChI=1S/C2H7N/c1-2-3/h2-3H2,1H3 and InChI=1S/C2H7N/c1-2-3/h2-3H2,1H3/p+1
        \item For the InChI notation, the protonation (h2-3H2,1H3) is identical in both cases an corresponds to ethylamine. For ethylammonium "p+1" indicates that an extra proton is added.
      \end{itemize}
    \\ \hline
       %Start of third eg
    \begin{minipage}{.2\textwidth}
      \includegraphics[width=\linewidth, height=23mm]{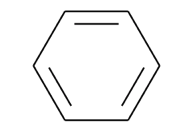}
      \center Benzene
    \end{minipage}
    &
    \begin{minipage}{.2\textwidth}
      \includegraphics[width=\linewidth, height=23mm]{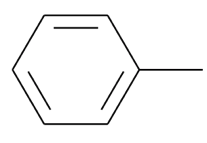}
      \center Toluene
    \end{minipage}
    & 
          \begin{itemize}
        \item SMILES: c1ccccc1 and Cc1ccccc1
        \item InChI=1S/C6H6/c1-2-4-6-5-3-1/h1-6H and InChI=1S/C7H8/c1-7-5-3-2-4-6-7/h2-6H,1H3
        \item For the SMILES notation, in the case of benzene atom 1 is connected to both atom 2 and atom 6, i.e. a ring is formed. "1" is a label and does not refer to atom number 1 (see toluene). A lower case "c" is used to indicate aromatic carbons, meaning they should be singly protonated.  For toluene, the methyl group is bonded to atom number 2, which is also bonded to atom number 7. For the InChI notation,  in the case of benzene aromaticity is inferred from the fact that all 6 carbon has 1 H atom (h1-6H).  For toluene, the methyl group is bonded to atom number 7, which is also bonded to atom number 6.
      \end{itemize}
    \\ \hline
  \end{tabular}
\end{table*}
SMILES and InChI are distinct notations. SMILES defines the bond types from which one can infer protonation, while InChI defines protonation from which one can infer the bond types. SMILES was designed to be read and written by humans and is therefore relatively straightforward to read, provided the user knows a few basic principles of the format. InChI, is comparatively less readable, is intended to ignore tautomeric form and is consistent. Table~\ref{tab:example_compounds} illustrates the distinction between SMILES and InChI representations. 

\subsection{Related Works}

There have been many efforts in the field of ML-assisted materials prediction and discovery~\cite{jha2018extracting, mannodi2016machine, jha2018elemnet, mozaffar2018data, pyzer2015learning, agrawal2016perspective, yang2019establishing, duvenaud2015convolutional, paul2019microstructure, goh2018smiles2vec, paul2018chemixnet}. 
For instance, Sajeev et al.~\cite{sajeev2013computational} developed a virtual library of semi-conductors and non semi-conductors of small organic compounds and applied various ML classifiers to predict the conductivity of a given compound. They used block descriptors as attributes which were generated based on E-Dragon Software~\cite{tetko2005virtual}. Riede et al.~\cite{riede2006datamining} used principal component analysis on a small data set of 62 organic solar cells which consisted of manufacturing parameters and measurement results on seven different substrates. Olivares-Amaya et al. ~\cite{olivares2011accelerated} employed Marvin code by ChemAxon~\cite{ChemAxon52:online} on a set of 50 training molecules compiled from the literature which provides a set of over 200 descriptors relevant for applications based on organic materials. Mannodi-Kanakkithodi et al.~\cite{mannodi2016machine} used kernel ridge regression (KRR) to transform the input fingerprint of a chemical compound into higher dimensional space to establish a linear relationship between the transformed fingerprint and the property of interest. They addressed the issue of accelerating polymer dielectrics design by extracting learning models from data generated by accurate state-of-the-art first-principles computations for polymers occupying a vital part of the chemical subspace. In their 2013 work, Kanal et al.~\cite{kanal2013efficient} focused on a screening pipeline which used a genetic algorithm for initial screening and multiple filtering stages for further refinement of the HOMO and LUMO properties of acceptors and donors. 
In the Harvard Energy Clean Project, Pyzer-Knapp et al.~\cite{pyzer2015learning} used a multi-layered perceptron (MLP) for predicting power conversion efficiency of organic photovoltaic materials from a sizable database of DFT-derived properties. 

Molecular graph convolutions~\cite{duvenaud2015convolutional} is an ML architecture for learning from undirected graphs, specifically from small molecules, and have become popular representations for molecular structures. It uses a simple encoding of the molecular graph—atoms, bonds, the distance between atoms that allow the model to take more advantage of information in the graph structure. More recently, Goh et al.~\cite{goh2018smiles2vec} developed an RNN neural network architecture SMILES2vec trained on SMILES for predicting chemical properties across different datasets. They did not explicitly encode information about SMILES grammar and rather allowed the RNN units to understand the intermediate features useful for property prediction implicitly. Their work also demonstrated that networks trained on line notations can perform better than other representations such as molecular graphs. Paul et al.~\cite{paul2018chemixnet} used a combination of fingerprints and SMILES to develop a deep neural network architecture called CheMixNet, and it outperformed SMILES2vec trained only on SMILES strings on the same datasets. 
\section{Method}
In this section, we present the source and target organic photovoltaic datasets used in our experiments, discuss the preprocessing of the SMILES and InChI strings and propound our methodology. 

\subsection{Datasets}
The source dataset for transfer learning is the Harvard CEP Dataset~\cite{pyzer2015learning, CEPData} which contains molecular structures and properties for 2.3 million candidate donor structures for organic photovoltaic cells. For a solar cell, the most important property is power conversion efficiency or the amount of electricity which can be generated due to the interaction of electron donors and acceptors, which are dependent on the HOMO values of the donor molecules. In this work, we considered the highest occupied molecular orbitals (HOMO) as the target property as it determines the power conversion efficiency of a solar cell according to the Scharber model~\cite{scharber2006design}.

The target dataset was the Harvard Organic Photovoltaic (HOPV) dataset~\cite{TheHarva43:online} which is a collection of photovoltaic measurements for a diverse set of 350 organic molecules generated by extensively searching the literature.  
Of these, experimental values were available for 243 molecules and calculated values using density functional theory (DFT) were available for 344 molecules.
In our experiments, the DFT-computed values in the HOPV dataset were reduced to 344 molecules after removing redundant isomeric samples~\cite{von2015fourier}. 
We used both the experimental and calculated datasets as target datasets for transfer learning.
The HOPV dataset contains density functional theory (DFT) calculations for four functionals B3LYP, BP86, PBE and M06 using the basis set def2-SVP ~\cite{weigend2005balanced}. 
We used B3LYP functional values as it is the most popular functional for HOMO value calculations. Further, HOMO values across all conformers were Boltzmann-weight averaged~\cite{barone2002determination}.

\subsection{Preprocessing}

For both the SMILES and InChI sequences,  one-hot encoding was performed separately to convert them into fixed length representations. 
The sequence lengths were calculated using the length of the longest SMILES and InChI sequences in the dataset respectively. To maintain a uniform sequence size, shorter strings were padded with zeros. 
Similar to SMILES2vec, vocabulary size was equal to the number of unique characters.
The sequence lengths are 82 and 162 respectively for SMILES and InChI input representations.

\begin{figure*}[]
\centering
\includegraphics[width = 1.0\textwidth, keepaspectratio=true]{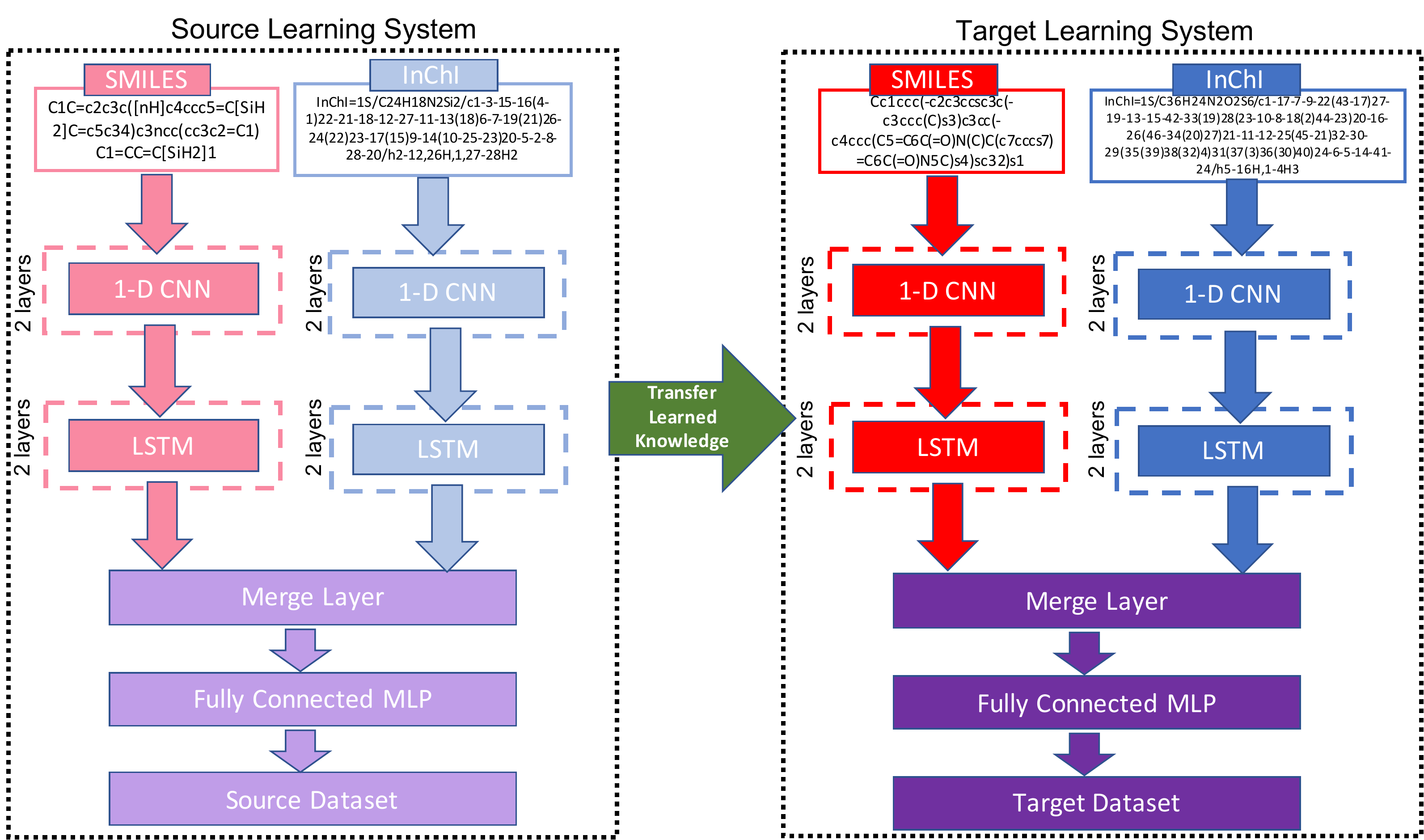}
\caption{The proposed SINet architecture for learning from the two text-based molecular representations - SMILES and InChI. The left side (represented by faded colors) represents the learning from the source dataset while the right side (represented by darker colors) represents the learning for the target dataset. For both the learning systems, the red branch represents the network for sequence modeling from SMILES while the blue branch represents the network for sequence modeling from InChI. The purple part represents the fully connected layers that learn the final output from the combination of features learned by the two network branches. We exemplify the SMILES and InChI with one representative example in this illustration from both the source as well as target datasets. }
\label{fig:model}
\end{figure*}

\subsection{SINet}

Figure~\ref{fig:model} illustrates the proposed approach for performing transfer learning from the Harvard CEP dataset to the two smaller target datasets in the HOPV dataset.
The deep neural network architecture used for the task consists of two branches for the two types of the input representations of SMILES and InChI.
The SMILES input vector has a length of 82 while the InChI input has 162 values.
Both branches have the same network configuration. Each branch is composed of a 1-D CNN followed by an LSTM network. The 1-D CNN is composed of two layers with 32 filters each; the filter size used in each layer is 3 and same padding for the inputs and output. The convolutional layers are followed by max pooling with a pool size of 2. There was no significant difference with other types of pooling and other pooling sizes. The output of the 1-D CNN is fed into the LSTM network which is composed of 2 layers having 64 units each.
Finally, the outputs from both branches are concatenated into the merge layer and fed into a fully connected network which is composed of a penultimate dense layer with 64 units and the final layer that gives the HOMO value as the regression output.
Since the network architecture leverages both SMILES and InChI molecular representations, we refer to it as SINet.

For transfer learning, first, we train a model on the source dataset of Harvard CEP from scratch (by initializing the model parameters from scratch before training). While being trained on the large dataset, the model learns a rich set of feature representations present in the large training data which is useful for making predictions in the source domain.
Next, for using transfer learning, we can follow one of the two techniques.
Either, the same trained model can be fine-tuned by training on the target dataset, or we can initialize a new model using the model parameters from the model trained on the source data and then fine-tune it on the target data. In this work, we use the first approach as the target dataset is very small, and we wanted to harness the source dataset as much as possible. In the case of transfer learning, rather than learning all the feature representations present in the input data from scratch, the model already knows the input data distribution from the source dataset and only fine-tunes its parameters to adapt to the target dataset.

\section{Experimental Results}
In this section, we present the experimental settings and results of the ensemble SINet architecture including the impact of transfer learning for performance gain on the smaller HOPV datasets. 

\subsection{Experimental Settings}
The models were implemented using Python and Keras~\cite{chollet2015keras} with TensorFlow~\cite{abadi2016tensorflow} as the backend. 
We used Adam as the optimization algorithm with a mini-batch size of 32. 
For generating the InChI fingerprints for the CEP dataset, we used RDKIT~\cite{landrum2006rdkit} library to generate InChI from the molecules. Scikit-Learn~\cite{pedregosa2011scikit} was used for data preprocessing and for evaluating the test set errors. 
All experiments are carried out using NVIDIA DIGITS DevBox with a Core i7-5930K 6 Core 3.5GHz desktop processor, 64GB DDR4 RAM, and 4 TITAN X GPUs with 12GB of memory per GPU. 
We performed extensive hyperparameter search as well as architecture search for SINet.
For our experiments, we used a learning rate of 0.001. 
We used the mean squared error (MSE) as the loss function and used the mean absolute \%  error (MAPE) as the performance metric.
Early stopping was used during training to avoid over-fitting.
For our experiments, we split each dataset into 70-20-10 ratio for training, test and validation sets; we used the same split for all experiments of each dataset.
Stratified shuffling was used to ensure that the distribution of HOMO values for all the 3 subsets was similar.

\begin{figure}[h]
\centering
\includegraphics[width = 0.48\textwidth, keepaspectratio=true]{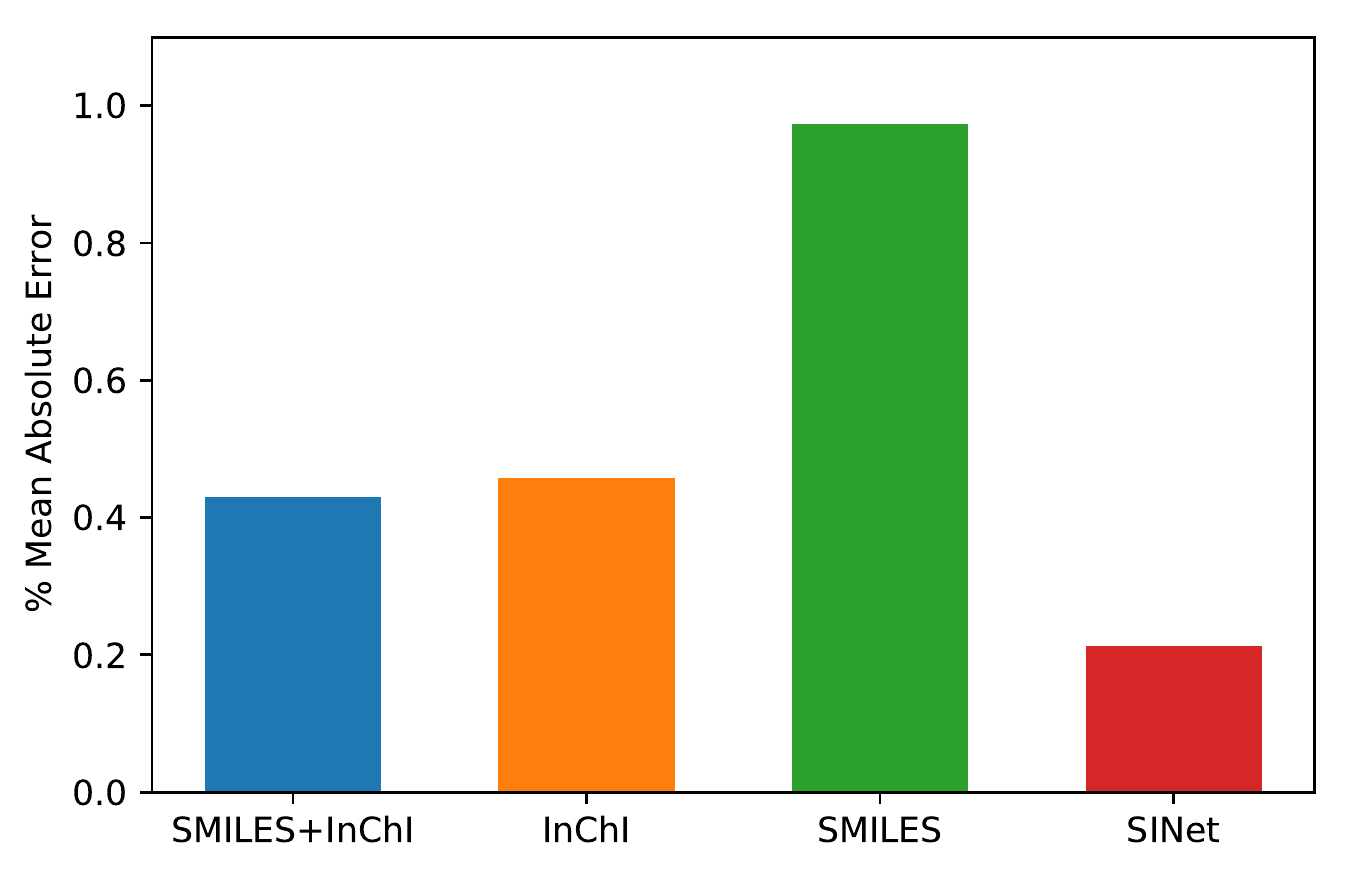}
\caption{Mean Absolute Error Percentage for the CEP Dataset (source dataset) }
\label{fig:cep_res}
\end{figure}

%\begin{figure*}[]
%\centering
%\subfigure[without transfer learning]{
%%\label{fig:sinet_exp_ntl}
%\includegraphics[width = 0.45\textwidth, keepaspectratio=true]{SINet_exp_ntl.pdf}}
%\qquad
%\subfigure[with transfer learning]{
%%\label{fig:sinet_exp_tl}
%\includegraphics[width = 0.45\textwidth, keepaspectratio=true]{SINet_exp_tl.pdf}}
%\caption{Impact of Transfer Learning on SINet for Experimental Dataset}
%\label{fig:exp_results}
%\end{figure*}

\begin{figure}[h]
\centering
\includegraphics[width = 0.48\textwidth, keepaspectratio=true]{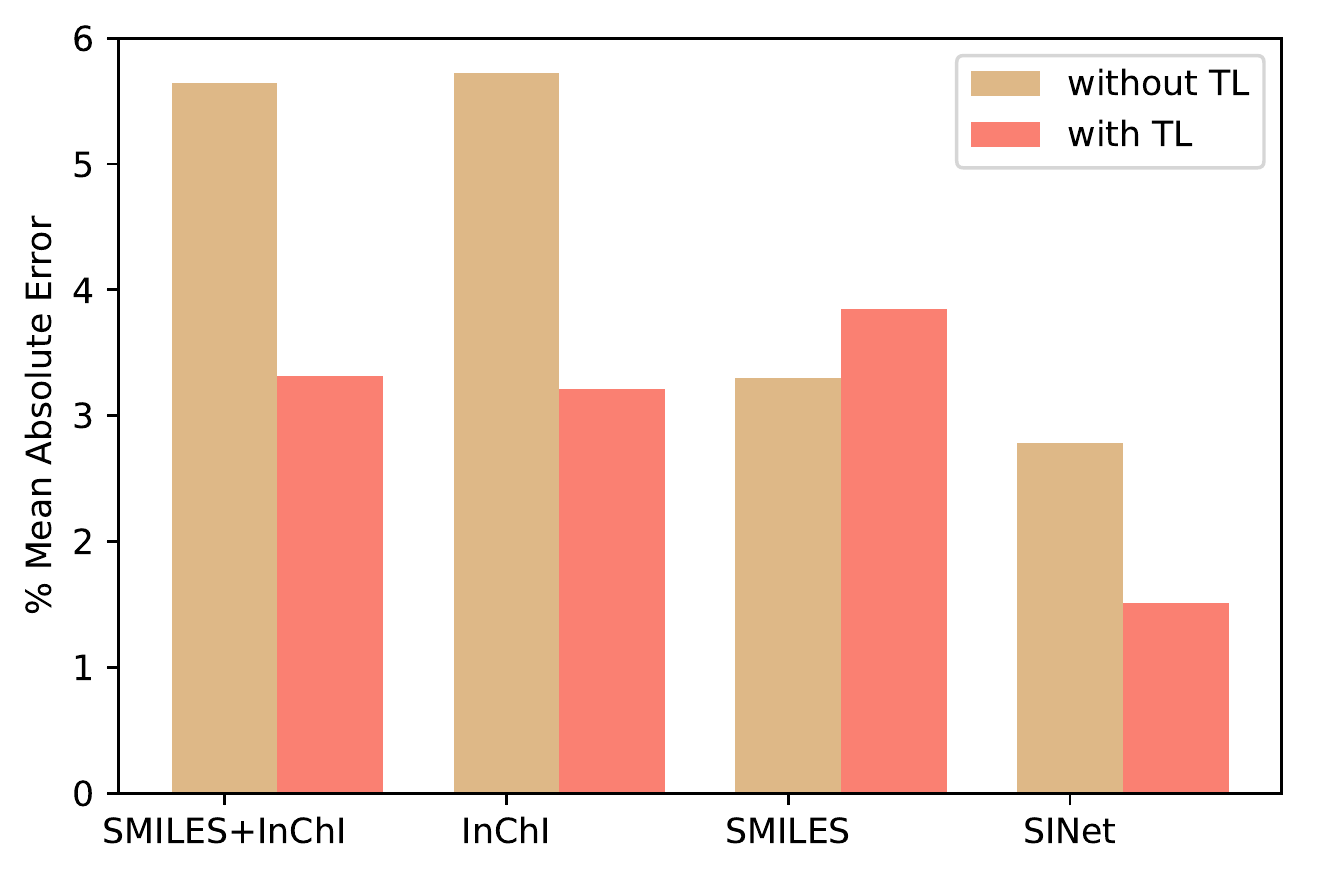}
\caption{Mean Absolute Error Percentage for the OPV Experimental Dataset without and with Transfer Learning (TL) }
\label{fig:exp_results}
\end{figure}

\begin{figure}[h]
\centering
\includegraphics[width = 0.48\textwidth, keepaspectratio=true]{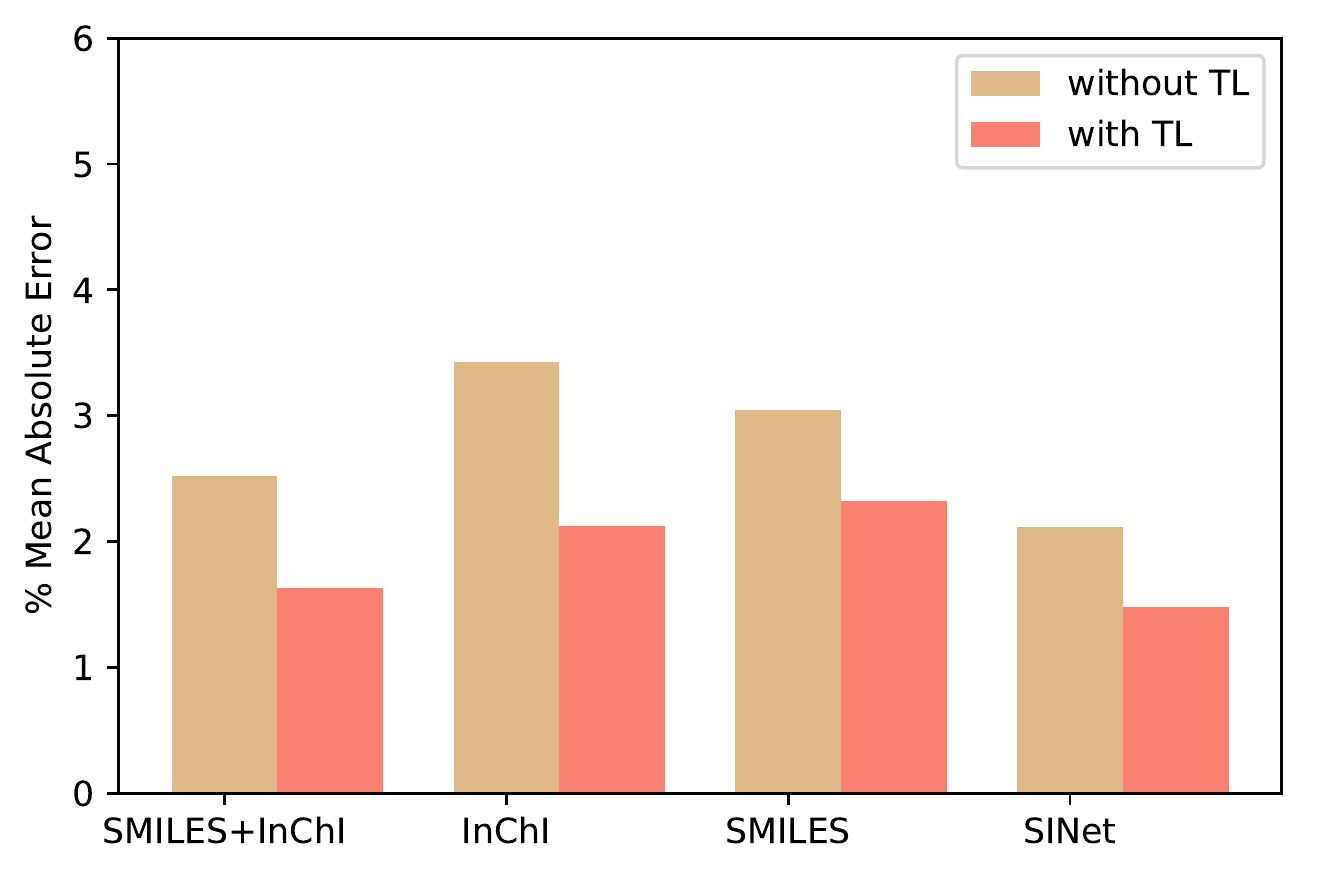}
\caption{Mean Absolute Error Percentage for the OPV DFT Dataset without and with Transfer Learning (TL) }
\label{fig:dft_results}
\end{figure}

%\begin{figure*}[]
%\centering
%\subfigure[without transfer learning]{
%%\label{fig:sinet_exp_ntl}
%\includegraphics[width = 0.45\textwidth, keepaspectratio=true]{SINet_calc_ntl.pdf}}
%\qquad
%\subfigure[with transfer learning]{
%%\label{fig:sinet_exp_tl}
%\includegraphics[width = 0.45\textwidth, keepaspectratio=true]{SINet_calc_tl.pdf}}
%\caption{Impact of Transfer Learning on SINet for DFT-Calculated Dataset}
%\label{fig:dft_results}
%\end{figure*}

\subsection{Impact of Leveraging SMILES \& InChi}
First, we explored the performance of using different types of input representations and their combinations on the source and the target datasets.
On the source dataset of Harvard CEP, we observe that the MAPE decreased to 0.213\% while using the SINet model as shown in Figure~\ref{fig:cep_res}. 
In contrast, while using the individual input representations with the individual branch of SINet, the MAPE values were 0.972\% and 0.457\% using SMILES and InChi.
This was also true for the target datasets of experimental and DFT-computed values from HOPV.
We conjecture the prediction mainly improves while leveraging multiple molecular representations because the two line notations- SMILES and InChI differ in the representation and detail; hence, the model can learn different feature representation from the two input representation, leading to better performance.

Furthermore, we experimented with simply combining the two input representations- SMILES and InChI into a single input vector (represented as SMILES + InChI in Figure~\ref{fig:cep_res}), before feeding them into a branch of SINet, the MAPE, in this case, was 0.430\% which is lower than while using single input.
However, there was no benefit from simply combining the two input representations into a single vector in the case of experimental target dataset.
We surmise that this could be because the two line notations encode different representations with varying lengths for different compounds, and a concatenation of the representations was not sufficient for learning both notations. Our results recommend that a better way to incorporate multiple input representations such as SMILES and InChI, in this case, is to design the deep neural network to have different model components to handle each of them before the learned features can be combined to make the final output as in the case of SINet.

In addition, we can observe the impact of training data size on the prediction performance; the prediction error of SINet on Harvard CEP dataset is significantly lower than the prediction error of SINet on the two other relatively smaller datasets. This also justifies the use of large dataset as the source dataset while using transfer learning to a smaller target dataset.

\subsection{Impact of Transfer Learning}
We also investigated the impact of transfer learning from the DFT-computed dataset of Harvard CEP to the relatively smaller DFT-computed and experimental datasets from HOPV.
For the experimental data having only 243 samples, the MAPE in case of SINet decreased significantly from 2.782\% to 1.513\% which is around half.
We observed similar changes when using just one input or their simple combination as shown in Figure~\ref{fig:exp_results}.
For the target dataset of DFT-computed dataset with 344 samples, the error for SINet decreased from 2.118\% to 1.478\% (in Figure~\ref{fig:dft_results}).
Such a significant drop in the MAPE for both our target datasets illustrates the efficacy of using transfer learning from large datasets when doing predictive modeling on smaller datasets with a lesser number of samples.
The experiments exhibit that when a model is trained on a large dataset (model parameters being initialized from a model trained on large source dataset), it already captures the required features from the dataset which makes it easy to learn the features present in target data from a similar domain on which it is fine-tuned.

\section{Conclusion and Future Works}
In this paper, we presented a novel approach of predictive modeling for HOMO values of donor molecules for the generation of OPV candidates by leveraging both large DFT-computed dataset and relatively smaller DFT-computed and experimental datasets using both types of input representation- SMILES and InChI, using the concept of transfer learning with deep neural networks.
For the source dataset, we leveraged the Harvard CEP dataset which contains millions of OPV candidates with the DFT-computed HOMO values. For the target dataset, we used the DFT-computed and experimental data from HOPV which contains relatively smaller data- 344 and 243, respectively.

Our results demonstrate significant benefit from the use of both types of input representations as well as from transfer learning from a larger dataset. It showcases that leveraging machine learning with computational and experimental chemistry can play an essential role in the expedition of a systematic design of high-efficiency OPV materials, and holds significant promise as a potential solution to future energy needs. The search process for the donor cells with high HOMO values can be made faster by leveraging transfer learning from a larger calculated dataset to a small well-curated experiment-theory calibrated dataset, and this exposes an exciting area in materials discovery, and in particular for solar cell technology. Further, as our approach is based on simple text representations, it is easier for chemists to explore adding or removing subgroups to the chemical compounds to explore the impact on power efficiency instead of performing elaborate experiments. 

As a future work, we believe Hierarchical Attention Networks (HANs)~\cite{yang2016hierarchical} that combine character level and word level sequences for text prediction can be harnessed to exploit the layer-based nomenclature of the InChI representation. Although we limited the scope of this current work for organic photovoltaic datasets, we believe that SINet would be helpful for the broad communities of cheminformatics and bioinformatics, as SMILES and InChI are popular representations for any type of organic molecule.

\section*{Acknowledgment}

This work is supported in part by the following grants: NIST award 70NANB14H012, NSF award CCF-1409601; DOE awards DE-SC0014330, DE-SC0019358.

\bibliographystyle{IEEEtran}

\bibliography{SINet.bib} 

\end{document}